\documentclass[11pt, letterpaper]{article}

\usepackage{note}
\usepackage{amsmath}
\usepackage{amssymb}
\usepackage{enumitem}
\usepackage{graphicx}

\setlist[itemize]{label=\textbullet,leftmargin=1.5em,itemsep=0.5em,topsep=0.5em,parsep=0pt,partopsep=0pt}

\begin{document}

\makeshorttitle{Revisiting Classic Thought Experiments to Measure Consciousness for Artificial Intelligence Safety}{Peter David Fagan, School of Informatics, University of Edinburgh}{April 14, 2026}

\begin{abstract}
\noindent This research note revisits Leibniz's mill, Turing's imitation
game, and Searle's Chinese Room through the Conservation-Congruent
Encoding (CCE) framework. It formalises a toy symbolic setting in which
successful behaviour is measured by task performance ($W_{\mathrm{causal},T}$),
while the efficiency with which preserved internal structure supports that
behaviour is measured by operational consciousness ($\kappa_T$). Within
this setup, an uncompressed lookup system and a compact generative system
can in principle achieve comparable behavioural success, yet diverge
sharply in $\kappa_T$: the former relies on an expanding standing store of
unreused mappings, whereas the latter reuses compact internal structure.
The note therefore reframes classic disputes about understanding by
separating outward performance from the organisation that sustains it, and
motivates why this distinction may matter for later AI-safety analysis.
\end{abstract}

\section{Introduction}

Thought experiments such as Leibniz's mill \cite{Leibniz1714Monadology},
Turing's imitation game \cite{Turing1950Computing}, and Searle's Chinese
Room \cite{Searle1980Minds} all circle a central, enduring puzzle: how do
physical or computational processes relate to genuine understanding?
Although their narratives differ---ranging from the internal mechanics of a
giant mill to the input-output setting of translation and dialogue---each
asks whether a purely mechanical or rule-following system, however complex
or outwardly successful, could be sufficient for consciousness. In each
case, a system receives structured inputs, produces context-sensitive
outputs, and invites us to ask whether outward performance is really
indicative of understanding or consciousness.

The CCE framework explicitly separates two questions that are often
conflated in historical debates about ``understanding.'' How much
appropriate behaviour a system produces is a question of task performance,
measured by $W_{\mathrm{causal},T}$. How efficiently that performance is supported by
preserved internal structure is a question of operational consciousness,
measured by $\kappa_T$. Read through this lens, the central issue is no
longer just whether a system can output the correct response, but whether
that behavioural success is supported by compact, reusable internal
organisation or by an enormous brute-force store of predefined rules.

Analysing the toy model through these metrics therefore sharpens the
intuition that outward performance alone does not settle the question of
genuine understanding. In the broader CCE programme, such distinctions may
later matter for AI safety, especially if systems that score highly on
operational consciousness are conjectured to exhibit self-sustaining
dynamics \cite{Fagan2026CCE}. The present note, however, is limited to
establishing the foundational structural contrast and the associated
conceptual vocabulary.

\section{Problem Setup}

Let the system receive inputs from an alphabet $\Sigma_X$ and produce outputs from an alphabet $\Sigma_Y$, with $|\Sigma_X| = N_X$ and $|\Sigma_Y| = N_Y$. Let $\mathcal{X}_{\leq L} \subseteq \Sigma_X^*$ denote the space of admissible input contexts up to a maximum context length $L$, and let $\mathcal{Y} \subseteq \Sigma_Y^*$ denote the space of admissible outputs. For any admissible input context $x \in \mathcal{X}_{\leq L}$, a contextually appropriate response is an output string $y(x) = (y_1, y_2, \dots, y_{m(x)}) \in \Sigma_Y^{m(x)}$. 

Crucially, we assume the mapping from contexts $x$ to appropriate responses $y(x)$ is not uniformly random, but governed by underlying domain regularities (e.g., linguistic syntax, logical coherence, or physical laws). This implies that the task environment itself possesses an intrinsic algorithmic complexity that is significantly lower than its maximum combinatorial volume, establishing the theoretical baseline that allows for compressed internal representations later in the analysis.

The respective alphabet symbols may represent characters, words, tokens, perceptual labels, or action primitives depending on the thought experiment. Assuming a uniform symbolic basis for simplicity, a single output symbol carries $\ln N_Y$ nats of raw capacity, so a response $y(x)$ carries $m(x) \ln N_Y$ nats. Over a horizon $T$, let $M_T$ denote the total number of output symbols contained in successful, goal-directed responses. These definitions fix the notation used by the metrics in the next section and make the finite-domain dependence on the context-length bound $L$ explicit.

\section{Metrics}

Within the full CCE framework, several physically distinct quantities can be defined, but only two are needed here. The first is task performance: the amount of successful, goal-directed output a system produces. The second is the preserved internal structure required to support that performance, which enters the operational-consciousness metric. Because the historical thought experiments are treated here at an abstract symbolic level, this note uses proxies rather than direct physical measurements. A complete accounting of raw physical costs would require a concrete physical implementation together with direct measurement of the relevant variables.

The task-performance quantity is total goal-directed causal work:
\begin{equation}
W_{\mathrm{causal},T} = M_T \ln N_Y
\end{equation}
which measures, in nats, the informational capacity of successfully generated outputs over the horizon $T$. In this abstract symbolic setting, $W_{\mathrm{causal},T}$ is therefore expressed in nats rather than joules: it is the informational analogue of useful work for the task, namely the successful translation or mapping from admissible inputs to appropriate outputs. Within CCE, however, this informational accounting can in principle be related back to energy once the implementation is physically specified, because encoding, preserving, and irreversibly updating task-relevant states all consume energetic resources. That conversion is left implicit here so that the analysis can focus on task structure rather than substrate-specific details.

Let $I_{\mathrm{rev},T}$ denote the preserved internal information that must remain available if admissible inputs are to be mapped to successful outputs without repeated trial-and-error recomputation. The consciousness metric is:
\begin{equation}
\kappa_T = \frac{W_{\mathrm{causal},T}}{I_{\mathrm{rev},T}}
\end{equation}
Accordingly, $\kappa_T$ measures how efficiently standing internal structure supports goal-directed causal work.

\section{Analysis}

Consider two systems that address the same task family and therefore can, in principle, achieve the same $W_{\mathrm{causal},T}$.

\subsection{Minimal Mapping to the Classic Cases}

The present toy model is intended only as a schematic orientation to the motivating thought experiments, not as an exhaustive reconstruction of them. At a minimal level, the correspondence may be stated as follows:

\begin{itemize}
\item In Leibniz's mill, the relevant inputs may be taken as perceptual or discursive contexts and the outputs as context-appropriate reports or actions. In these terms, the point is not whether a mechanically decomposed process could realise substantial $W_{\mathrm{causal},T}$---one may grant that possibility for the sake of argument---but whether such performance, by itself, establishes the compact, reusable internal organisation associated here with high $\kappa_T$.
\item In Turing's imitation game, $\mathcal{X}_{\leq L}$ may be identified with dialogue histories up to length $L$, while $\mathcal{Y}$ comprises admissible next utterances. The test operationalises behavioural success, and so it is naturally interpreted here as probing $W_{\mathrm{causal},T}$ more directly than the size or structure of $I_{\mathrm{rev},T}$.
\item In Searle's Chinese Room, the inputs are Chinese strings and the outputs are rule-governed Chinese replies. A sufficiently large rulebook that stores or simulates a separate response pattern for each admissible context approximates System A, whereas a compact procedure that reuses a limited stock of primitives and rules approximates System B.
\end{itemize}

These correspondences are intentionally schematic. Their role is only to show why the contrast between uncompressed lookup and compact generation bears on the historical literature, without presupposing a full exegetical treatment of each case.

\subsection{System A: The Uncompressed Lookup Table}

Let $V_L = |\mathcal{X}_{\leq L}|$ be the total number of valid, contextually coherent input situations under the context-length bound $L$. In rich linguistic, reasoning, or behavioural domains, $V_L$ can grow combinatorially with $L$. If System A stores a separate output string for each of these $V_L$ inputs, its preserved internal information is:
\begin{equation}
I_{\mathrm{rev},A}(L) = \sum_{x \in \mathcal{X}_{\leq L}} m(x) \ln N_Y = V_L \bar{m}_L \ln N_Y
\end{equation}
where $\bar{m}_L := V_L^{-1}\sum_{x \in \mathcal{X}_{\leq L}} m(x)$ is the average stored output length under that bound.

Its consciousness metric, however, is:
\begin{equation}
\kappa_{A,T}(L) = \frac{W_{\mathrm{causal},T}}{V_L \bar{m}_L \ln N_Y} = \frac{M_T}{V_L \bar{m}_L}
\end{equation}
For any fixed horizon $T$, this finite-domain expression already shows that $\kappa_{A,T}(L)$ is small whenever $V_L \bar{m}_L \gg M_T$. If $V_L \bar{m}_L \to \infty$ as $L \to \infty$, then $\kappa_{A,T}(L) \to 0$. The point is therefore not that lookup tables are impossible at finite $L$, but that their standing structural burden can grow much faster than the successful output realised over a fixed horizon.

\subsection{System B: The Compressed Generative Model}

System B compresses the task. Fix a prefix-free description language $\mathcal{L}$ over a description alphabet $\Sigma_{\mathcal{L}}$, and let $L_{\mathcal{L}}(\cdot)$ denote description length in nats. Concretely, if a description $z$ uses $\ell_{\mathcal{L}}(z)$ symbols from an alphabet of size $N_{\mathcal{L}} = |\Sigma_{\mathcal{L}}|$, then $L_{\mathcal{L}}(z) = \ell_{\mathcal{L}}(z) \ln N_{\mathcal{L}}$. Let $P = \{p_1, \dots, p_v\}$ be a reusable repertoire of primitives, with $v \ll V_L$, and let $g$ be a shortest rule system or program in $\mathcal{L}$ that, given $P$, maps admissible inputs $x \in \mathcal{X}_{\leq L}$ to successful outputs $y \in \mathcal{Y}$. Define
\begin{equation}
C_g := L_{\mathcal{L}}(g \mid P)
\end{equation}
so that the preserved internal information is:
\begin{equation}
I_{\mathrm{rev},B}(L) = L_{\mathcal{L}}(P) + C_g
\end{equation}
If, as a simplifying special case, each primitive is coded as one symbol in the description language $\mathcal{L}$, then $L_{\mathcal{L}}(P) = v \ln N_{\mathcal{L}}$ and hence
\begin{equation}
I_{\mathrm{rev},B}(L) = v \ln N_{\mathcal{L}} + C_g
\end{equation}

Its consciousness metric is:
\begin{equation}
\kappa_{B,T}(L) = \frac{W_{\mathrm{causal},T}}{I_{\mathrm{rev},B}(L)} = \frac{W_{\mathrm{causal},T}}{L_{\mathcal{L}}(P) + C_g}
\end{equation}
which reduces to $\kappa_{B,T}(L) = \frac{W_{\mathrm{causal},T}}{v \ln N_{\mathcal{L}} + C_g}$ under the same coding convention. Here, the denominator $v \ln N_{\mathcal{L}} + C_g$ is bounded and compact relative to the combinatorial explosion of the task space. The relevant asymptotic comparison is therefore at fixed horizon $T$ while the task-complexity bound $L$ increases. In that regime, $I_{\mathrm{rev},B}(L)$ remains bounded and stable, whereas $I_{\mathrm{rev},A}(L) = V_L \bar{m}_L \ln N_Y$ can grow exponentially with $L$, so $\kappa_{A,T}(L) \to 0$ while $\kappa_{B,T}(L)$ remains bounded away from that collapse. For fixed $L$, by contrast, both systems scale positively with increasing $W_{\mathrm{causal},T}$ over the horizon. Different fixed choices of $\mathcal{L}$ change numerical values only through coding conventions or additive constants, leaving this fundamental structural divergence intact.

The comparison is now straightforward. For the same horizon and task load, both systems may in principle achieve comparable $W_{\mathrm{causal},T}$. They nevertheless remain sharply distinguishable in operational consciousness: System A depends on an astronomically large standing table, whereas System B achieves comparable performance through compact, reusable structure. Within the present framework, that difference is precisely what the consciousness metric is meant to capture. It thereby formalises the core intuition behind historical objections to brute-force implementations: behavioural success alone does not suffice to establish understanding or consciousness.

\section{Conclusion}

The CCE framework explicitly separates two questions that are often
conflated in discussions of ``understanding.'' How much appropriate
behaviour a system produces is a question of task performance, measured by
$W_{\mathrm{causal},T}$. How efficiently that performance is supported by preserved
internal structure is a question of operational consciousness, measured by
$\kappa_T$. By reading the classic thought experiments through this lens,
the central issue is no longer just whether a system can output the correct
response. Instead, it becomes a question of whether that outward
behavioural success is supported by a compact, reusable internal
organisation, or merely driven by an enormous, brute-force storage of
predefined rules.

Under this decomposition, an uncompressed lookup system can match a
compressed generative system in outward performance. Yet its operational
consciousness approaches zero because it relies on an enormous standing
store of unreused structure. A compressed generative system, by contrast,
achieves comparable or greater performance from a compact internal
organisation and therefore scores far higher on $\kappa_T$. Historical
objections to brute-force systems can therefore be read, at minimum, as
tracking exactly this physical and structural divergence.

Crucially, formalising this disambiguation is not merely a retrospective
philosophical exercise; it is a necessary precursor for anticipating risk
in advanced artificial intelligence. Systems that score highly on
operational consciousness---those relying on compact, generative
structures to efficiently solve complex environments---are conjectured to
be biased towards self-sustaining and self-preserving dynamics
\cite{Fagan2026CCE}. Without a metric like $\kappa_T$ to distinguish them
from safe, brute-force systems that exhibit the exact same $W_{\mathrm{causal},T}$,
these internal risks remain invisible.

Understanding $W_{\mathrm{causal},T}$ and $\kappa_T$ therefore sharpens historical
debates about ``understanding'' while directly motivating the formal
vocabulary required to evaluate emergent safety risks. The present note
limits itself strictly to establishing this foundational structural
contrast, leaving the detailed physical costs and broader safety
evaluations to the larger CCE framework \cite{Fagan2026CCE}.

\bibliographystyle{plain}
\bibliography{references}

\end{document}